\pdfoutput=1
\documentclass{article}
\usepackage{spconf,amsmath,graphicx,booktabs}
\usepackage[hidelinks]{hyperref}

\title{Hand-Aware Transition Modeling for Bimanual\\Procedural Anomaly Detection}
\name{\shortstack{Di Wen\textsuperscript{1,*}, Jimmy Weissert\textsuperscript{1,*}, Luc Maria Scherrer\textsuperscript{1,*}, Cedric Z\"{o}llner\textsuperscript{1,*}, Kailun Yang\textsuperscript{2},\\ Ruiping Liu\textsuperscript{1}, Yufan Chen\textsuperscript{1}, Jiale Wei\textsuperscript{1}, Junwei Zheng\textsuperscript{1}, Kunyu Peng\textsuperscript{1,\textdagger}}\thanks{\textsuperscript{*}Equal contribution (email: di.wen@kit.edu).\ \textsuperscript{\textdagger}Corresponding author (email: kunyu.peng@kit.edu).}}
\address{\textsuperscript{1}Karlsruhe Institute of Technology, Karlsruhe 76131, Germany\\ \textsuperscript{2}Hunan University, Changsha 410012, China}

\begin{document}
\maketitle

\begin{abstract}
Procedural anomaly detection in bimanual assembly requires judging each hand action against the execution so far. A corrective action may look unusual in isolation, while a visually plausible action can violate the order of the procedure. We present HACT, a transition model over predicted per-hand events. A role-preserving history keeps the concurrent responsibilities of both hands, and a marked temporal point process assigns each observed transition a semantic and temporal surprisal. A supervised evidence head and a two-state filter convert these surprisals into per-hand anomaly posteriors. A recovery-aware protocol on predicted events and participant-disjoint folds reports the recovery false-positive rate at an operating point selected on validation participants. On two bimanual power-tool procedures HACT has the highest AUPRC and F1 among the compared methods and the fewest recovery alarms. Applied without retraining to a different assembly order of the same product, it retains the highest AUPRC and F1. The source code is available at \url{https://github.com/Kratos-Wen/HACT}.
\end{abstract}


\section{Introduction}
\label{sec:intro}
Industrial assembly is performed with two hands whose actions overlap in time and differ in role.
A procedural anomaly, such as a wrong tool, a wrong part, a step out of order, or a mishandled component, is defined relative to the progress made so far and to the alternatives the procedure allows.
Detectors judge each new action against a representation of the execution: a prototype, the anticipated next step, a task graph, or a recurrent state~\cite{lee2024error,flaborea2024prego,huang2025modeling,patsch2025mistsense}.
Corrections are frequent in procedural work and are annotated as such in Assembly101 and IMPACT~\cite{sener2022assembly101,wen2026impact}.
A detector dominated by local evidence may therefore assign a high anomaly score to the correction, because its validity is explained by the preceding mistake rather than by appearance alone.

\begin{figure}[t]
\centering
\includegraphics[width=\columnwidth]{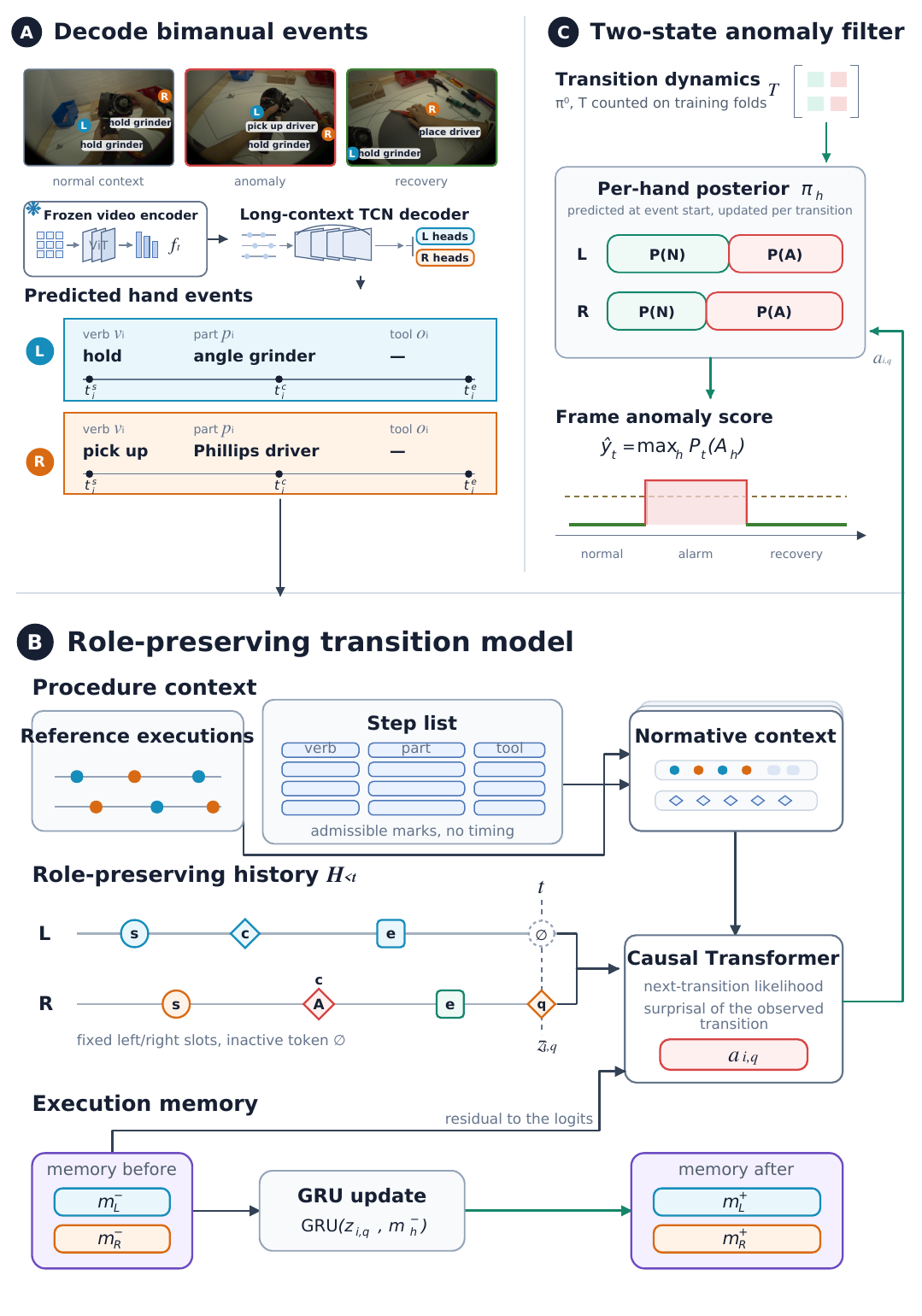}
\caption{HACT. Predicted hand events enter the role-preserving history $H_{<t}$; the surprisal of each transition gives the phase evidence $\widetilde a_{i,q}$, calibrated to $a_{i,q}$ for the per-hand posterior.}
\label{fig:method}
\vskip-1ex
\end{figure}

HACT (Hand-Aware Causal Transition) models the execution as a causal stream of structured transitions of the two hands, decoded from the video, and scores each transition by how unexpected it is under the joint history of both hands and the continuations the procedure admits (Fig.~\ref{fig:method}).

Our contributions are threefold. (i) A role-preserving bimanual transition model: a marked temporal point process over the events of both hands in fixed role slots, whose semantic and temporal surprisals form the anomaly evidence. (ii) A fully predicted, participant-disjoint protocol for recovery-aware detection that reports the recovery false-positive rate at a validation-selected operating point on all and on covered recovery frames. (iii) Controlled comparisons on two bimanual power-tool procedures and a transfer without retraining to another assembly order of the same product, in which HACT has the highest AUPRC and F1 and the fewest recovery alarms.

\section{Related Work}
\label{sec:related}
\noindent\textbf{Procedural mistake detection.}
EgoPER compares each segment with prototypes of normal actions~\cite{lee2024error}; PREGO and TI-PREGO raise an alarm when the recognized action disagrees with the anticipated one~\cite{flaborea2024prego,plini2026tiprego}; task graphs admit alternative valid orders, learned differentiably, generalized, or expanded into several normal representations of the next actions~\cite{seminara2024differentiable,lee2025error,huang2025modeling}.
AEM models the effect of an action instead of the action~\cite{guo2026procedural}, MistSense separates procedural from execution mistakes online~\cite{patsch2025mistsense}, and ESTANet uses the inconsistency of several online predictors~\cite{lee2026estanet}.
Vision--language models detect mistakes zero-shot~\cite{ozsoy2026unreasonable}, after post-training~\cite{spurio2026posttraining}, or with explanations~\cite{lee2026axg}; later work attributes the mistake~\cite{li2026mistake} or balances its long tail~\cite{han2026understanding}.
Few of these model the two hands as separate actors of one execution.

\noindent\textbf{Benchmarks and event models.}
Assembly101 introduced step recognition with natural corrections~\cite{sener2022assembly101}, ATTACH labels hand-specific industrial actions~\cite{aganian2023attach}, IndustReal, EgoPER, CaptainCook4D and HoloAssist annotate execution errors at the step level~\cite{schoonbeek2024industreal,lee2024error,peddi2024captaincook4d,wang2023holoassist}, and PIE-V injects errors and corrections into clean procedures~\cite{loginova2026how}.
IMPACT provides decoupled bimanual hand--object events with anomaly and recovery labels for a power-tool procedure~\cite{wen2026impact,zhang2026impact}.
Recurrent and Transformer intensity models describe event streams~\cite{du2016recurrent,zuo2020transformer}, and goodness-of-fit tests of a fitted point process detect anomalous sequences~\cite{shchur2021detecting}; we score single transitions by their surprisal under a marked process over both hands.

\section{Method}
\label{sec:method}
\noindent\textbf{Problem formulation.}
Given an egocentric video $V=\{I_t\}_{t=1}^{T}$ and a procedure context $\mathcal C$, the task is a frame-level anomaly score $\widehat y_t\in[0,1]$.
The setting is offline detection on a recorded execution: the event decoder reads the whole recording, a frame is scored once the event containing it has been decoded, and the transition model is causal in the event stream.
The context is two correct executions and a step list; the executions enter as decoded event streams, the step list as a set of admissible marks without order, and both are fixed before cross-validation.

\noindent\textbf{Bimanual event decoding.}
A long-context temporal convolutional network maps the frozen frame features, without a hand or object detector, to per-frame phase, verb, part and tool logits for each hand, one linear head per factor and hand on a shared trunk.
A Viterbi pass over the whole recording enforces the phase order idle, approach, interaction, idle without learned penalties.
Consecutive decoded states form events $e_i=(h_i,\mathbf t_i,v_i,p_i,o_i)$ with hand $h_i\in\{L,R\}$, predicted start, onset, and end frames $\mathbf t_i=(t_i^s,t_i^c,t_i^e)$, and verb, part and tool labels $(v_i,p_i,o_i)$, the argmax of the logits averaged over the frames of the event.

\noindent\textbf{Role-preserving transition model.}
The hands act concurrently in different roles, one stabilizing while the other operates, and pooling concurrent tokens into one vector discards which hand did what.
Each event yields three transitions $(i,q)$, $q\in\{s,c,e\}$, embedded as
\begin{equation}
\begin{aligned}
 z_{i,q}={}&W_x x_{i,q}+E_h(h_i)+E_q(q)+E_v(v_i)+E_p(p_i)\\
          &+E_o(o_i)+E_\sigma(\sigma_i)+\phi\!\left(\log(1+\Delta_{i,q})\right),
\end{aligned}
\end{equation}
where $x_{i,q}$ concatenates the boundary and phase-pooled decoder states, $\sigma_i$ marks query, reference, or step-list input, and $\phi$ embeds the elapsed time $\Delta_{i,q}$ since the previous transition of either hand; a step-list token carries only its mark embeddings, with a neutral hand and no decoder state or elapsed time.
Transitions of both hands that fall on the same decoded frame form one simultaneous group: tokens of the same hand in a group are averaged, left and right occupy fixed slots, and a hand without a transition in the group is represented by one learned inactive token.
A linear projection of the two slots $(u_L,u_R)$, their difference $u_L-u_R$, and their product $u_L\odot u_R$ forms one bimanual group token, so that relative and joint activity of the hands are linearly available.
A causal Transformer returns the context $H_{<t}$ before the current group, so every prediction enters the history but never sees itself, and cross-attention from $H_{<t}$ retrieves normative context from the decoded reference executions and the step-list tokens.
Conditioned on history and context, one shared representation feeds four linear mark heads for the phase transition, verb, part, and tool of the next transition and a rate head with one non-negative rate $\lambda_L,\lambda_R$ per hand~\cite{du2016recurrent,zuo2020transformer}.
Let $\Lambda=\lambda_L+\lambda_R$. The waiting time to the next transition of either hand is exponential with total rate $\Lambda$, and the next acting hand follows $P(h)=\lambda_h/\Lambda$. Together with the four categorical mark heads this gives the negative log-likelihood of the observed transition of hand $h$ after $\Delta$ seconds,
\begin{equation}
\mathcal L_{\mathrm{tr}}=-\log\lambda_h+\Lambda\Delta-\textstyle\sum_{r\in\{q,v,p,o\}}\log p_r(r_i\mid H_{<t},\mathcal C);
\end{equation}
hand identity and waiting time share one cause-specific temporal term, and the transitions of a simultaneous group are scored one by one against the same $H_{<t}$.
We decompose the transition likelihood into the hand, mark, waiting-time, and survival ($\Lambda\Delta$) surprisals and concatenate them with a visual residual, one minus the cosine similarity between $z_{i,q}$ and the attended reference context; a small head maps this evidence vector to binary logits and to logits over the annotated anomaly types.
Per-hand execution memories $(m_L,m_R)$ are initialized from the encoded normative context and written at every transition of the hand, $m_h^{+}=\mathrm{GRU}(z_{i,q},m_h^{-})$. The two memories are fused in role order, $(m_h,m_{\bar h})$, with the same slot, difference, and product terms as the history representation. A zero-initialized adapter compares $z_{i,q}$ with this memory context through their normalized difference and adds the resulting residual to the binary logits.

\noindent\textbf{Two-state anomaly filter.}
Binary logits are temperature-scaled, giving the phase evidence $a_{i,q}$.
For each hand in each video we maintain a posterior $\boldsymbol\pi_h$ over $\{N,A\}$ with initial distribution $\boldsymbol\pi^{0}$ and transition matrix $\mathbf T$ counted from the anomaly-status sequences of the training recordings ($T_{jk}=n_{jk}/\sum_{k'}n_{jk'}$, unsmoothed).
The transition matrix is applied once, at the start transition of an event, $\boldsymbol\pi_h^{-}=\mathbf T^{\top}\boldsymbol\pi_h^{+}$; the onset and end transitions inherit the preceding posterior; every transition contributes its phase evidence through the update $\boldsymbol\pi_h^{+}\propto\boldsymbol\pi_h^{-}\odot[1-a_{i,q},\,a_{i,q}]$.
Because the evidence head is trained under a balanced class prior, the calibrated odds $a_{i,q}/(1-a_{i,q})$ serve as soft evidence, and the class prior enters the filter once, through $\boldsymbol\pi^{0}$ and $\mathbf T$.
$P(A_h)$ is held over the decoded phase interval, $\widehat y_t=\max_h P_t(A_h)$, and frames outside the predicted support receive zero.

\noindent\textbf{Learning objective.}
The modules are trained in dependency order with the earlier stages frozen: event decoding, transition likelihood on correct transitions, anomaly evidence, and execution memory; the staging keeps the scarce anomaly supervision from reshaping the representation learned from normal transitions.
The transition model and the evidence head see both the annotated event streams and the streams produced by the trained decoder, so decoder errors are present during training.

\section{Experiments}
\label{sec:exp}
\begin{table}[t]
\centering
\caption{Fully predicted comparison. R-FPR: recovery false-positive rate at the validation-selected recall-.3 threshold on all recovery frames, R-FPR$_{\mathrm{cov}}$: on covered recovery frames.}
\label{tab:main}
\resizebox{\columnwidth}{!}{%
\begin{tabular}{@{}lcccc@{}}
\toprule
Method & AUPRC $\uparrow$ & F1 $\uparrow$ & R-FPR $\downarrow$ & R-FPR$_{\mathrm{cov}}$ $\downarrow$ \\
\midrule
\multicolumn{5}{@{}l}{\emph{IMPACT-ego Reassembly (public), 10.5\% anomalous frames}} \\
Causal TCN~\cite{lea2017temporal} & .095 & .224 & .284 & .284 \\
MistSense~\cite{patsch2025mistsense} & .129 & .224 & .325 & .325 \\
EgoPER-V~\cite{lee2024error} & .115 & .182 & .448 & .450 \\
HACT & \textbf{.160} & \textbf{.252} & \textbf{.225} & \textbf{.281} \\
\midrule
\multicolumn{5}{@{}l}{\emph{Second bimanual set, 9.0\% anomalous frames}} \\
Causal TCN~\cite{lea2017temporal} & .112 & .193 & .296 & .296 \\
MistSense~\cite{patsch2025mistsense} & .116 & .191 & .211 & .211 \\
EgoPER-V~\cite{lee2024error} & .097 & .151 & .286 & .286 \\
HACT & \textbf{.152} & \textbf{.246} & \textbf{.083} & \textbf{.096} \\
\bottomrule
\end{tabular}
}
\vskip-1ex
\end{table}
\noindent\textbf{Datasets.}
IMPACT-ego v1.1~\cite{wen2026impact} records an angle-grinder procedure from the head-mounted camera.
We use its Reassembly work package: 41 executions by 12 evaluation participants, 184k frames of which 10.5\% are anomalous.
The two recordings of the remaining participant, excluded from every fold, are the reference executions, and the step list is derived from them.
Anomalous segments occur in unseen context far more often than normal ones: with leave-one-recording-out over its annotation, the other hand's concurrent action is an unseen combination for 14 to 20\% of the anomalous segments against 6\% of the normal ones, and the same-hand predecessor for 29 to 38\% against 15\%.
We additionally evaluate a second bimanual collection recorded on a modified angle-grinder disassembly setup: 31 recordings by 13 further participants, 116,957 frames, 9.0\% anomalous, 183 anomaly events and 72 recovery events; its references carry no event annotation, so its step list is the written procedure specification. The collection will be released with the evaluation code.
Both sets use five participant-disjoint folds; every recording is tested once and AUPRC is pooled over all out-of-fold frames.
IMPACT-ego also provides a second reassembly work package on the same angle grinder (Reassembly~B): the same vocabulary in a different assembly order, in which one in five correct step transitions never occurs in A; 10 executions by 10 of the evaluation participants, 52.6k frames of which 18.7\% are anomalous, 180 anomaly events and 33 recoveries.

\noindent\textbf{Implementation details.}
Frames are encoded by a frozen encoder at 15 frames per second: the 1,408-D VideoMAE~V2 features shipped with the public benchmark~\cite{wang2023videomaev2}, and 1,280-D DINOv3 ViT-H+/16 features on the second set~\cite{simeoni2025dinov3}.
The decoder is a 128-D dilated temporal convolutional trunk with a 511-frame receptive field (34\,s); the transition model has one Transformer layer, 128-D states, four heads and a 64-D evidence head; decoder and transition model together have 1.49M parameters beyond the encoder.
All stages use AdamW with weight decay $10^{-3}$. The decoder uses class-count-balanced softmax losses per factor and hand; the transition model minimizes the factorized transition negative log-likelihood with a cross-hand auxiliary weighted 0.25; the evidence head uses a class-count-balanced binary loss with an anomaly-type term weighted 0.25, recovery frames weighted 2.0 and hard negatives 1.5.
Learning rates are $10^{-3}$ for the decoder and $3{\cdot}10^{-4}$ afterwards; early stopping, model selection by event-level AUPRC, temperature and thresholds all use the validation participants.

\noindent\textbf{Protocol and metrics.}
Every learned method receives the same cached query features, no annotation of the query, and predicts its own support and procedural representation; scores are rasterized over that support, hands are combined by maximum, and uncovered frames receive zero.
For the recovery false-positive rate (R-FPR), the fraction of annotated recovery frames that are flagged, each fold selects on its validation participants the threshold at which recall reaches .3 and applies it once to the test recordings; the achieved test recall is reported with the rate.
Uncovered frames can never be flagged, so we also report the decoded-support coverage per frame class and the recovery rate on covered recovery frames only (R-FPR$_{\mathrm{cov}}$).
For Reassembly~B the five fold models of A are applied unchanged: each B recording is scored by the fold model that excluded its participant, and F1 uses the threshold selected on that fold's A validation participants. B has no validation participants, so no recovery rate is reported.

\noindent\textbf{Baselines.}
PREGO~\cite{flaborea2024prego} uses its released MiniROAD recognizer, a frozen Qwen3-30B anticipator with training-fold context, and its native mismatch decision, which yields a binary decision and no ranking.
EgoPER-V~\cite{lee2024error} is the official ActionFormer/normal-prototype pipeline on the shared features with the object-graph branch off.
MistSense~\cite{patsch2025mistsense} is its RGB Video-Q-Former classification path on the shared features, and the causal TCN~\cite{lea2017temporal} is a dilated temporal convolutional classifier over the same features; both are supervised with frame-level anomaly labels and produce a dense score for every frame without a procedure representation; all four use the same folds and selection rule.
Zero-shot Qwen2.5-VL-7B~\cite{bai2025qwen25vl} follows~\cite{ozsoy2026unreasonable} and is omitted from the table. Step-level task-graph methods~\cite{huang2025modeling,lee2025error,guo2026procedural} require an alignment between their step vocabulary and the repeating verb--part--tool event stream and are not compared.

\noindent\textbf{Comparison with the state of the art.}
Table~\ref{tab:main} shows that HACT has the highest AUPRC and F1 on both sets, .031 AUPRC over MistSense on Reassembly and .036 on the second set; PREGO reaches F1 .220 and .130 with its binary decision, and zero-shot Qwen2.5-VL-7B ranks at the random level with AUPRC .105 and .089.
Unlike the dense classifiers and the prototype baseline, HACT scores each decoded transition under an explicit history of both hands, and unlike PREGO it models a distribution over structured transitions instead of the agreement between recognition and anticipation; removing that history on the second set lowers AUPRC from .152 to .116 (Table~\ref{tab:ablation}).

\noindent\textbf{Recovery alarms.}
At the recall-.3 operating point HACT has the lowest rate on all recovery frames on both sets: .225 at a test recall of .36 against .325 at .37 for MistSense on Reassembly, and .083 at .34 against .211 at .37 on the second set. The ordering persists at recall .4 and .5 (.351 and .374 against .452 and .563; .185 and .199 against .355 and .448).
HACT's decoder covers .91/.80/.80 of the anomalous/normal/recovery frames on Reassembly (.94/.83/.86 on the second set), so recovery frames are not skipped preferentially, and on covered frames only HACT keeps the lowest rate on both sets (.281 against .284 for the causal TCN; .096 against .211).
The advantage does not come from the recovery-frame weight of the evidence loss alone: trained without it, HACT flags .216 and .196 of the recovery frames, still below every baseline.

\begin{table}[t]
\centering
\caption{Transfer to Reassembly B (18.7\% anomalous frames): the Reassembly A fold models applied unchanged.}
\label{tab:transfer}
\begin{tabular}{@{}lcc@{}}
\toprule
Method & AUPRC $\uparrow$ & F1 $\uparrow$ \\
\midrule
Causal TCN~\cite{lea2017temporal} & .153 & .343 \\
MistSense~\cite{patsch2025mistsense} & .218 & .347 \\
HACT & \textbf{.238} & \textbf{.358} \\
\bottomrule
\end{tabular}
\vskip-1ex
\end{table}
\noindent\textbf{Transfer to a different assembly order of the same product.}
Applied unchanged to Reassembly~B (Table~\ref{tab:transfer}) with the Reassembly-A procedure context, HACT has the highest AUPRC and F1, .238 and .358 against .218 and .347 for MistSense.

\begin{table}[t]
\centering
\caption{Ablations on the second bimanual set; R-FPR at the validation-selected recall-.3 threshold.}
\label{tab:ablation}
\begin{tabular}{@{}lccc@{}}
\toprule
Variant & AUPRC $\uparrow$ & F1 $\uparrow$ & R-FPR $\downarrow$ \\
\midrule
HACT & .152 & .246 & .083 \\
\quad no role-preserving history & .116 & .212 & .103 \\
\quad no interaction terms & .153 & .194 & .217 \\
\quad no memory & .143 & .213 & .207 \\
\quad gated write & .143 & .208 & .233 \\
\quad visual-only & .128 & .214 & .129 \\
\bottomrule
\end{tabular}
\vskip-1ex
\end{table}
\noindent\textbf{Ablations.}
The ablations (Table~\ref{tab:ablation}) separate the roles of history, hand interaction, and memory. Removing the role-preserving history causes the largest AUPRC drop, so the procedural context is the main source of the ranking. Removing the interaction terms leaves AUPRC unchanged but lowers F1 and raises the recovery rate from .083 to .217, so the explicit interaction of the two hands acts mainly on the recovery behavior; on Reassembly the same removal raises it from .225 to .383. The execution memory supports both ranking and recovery, and the gated write of our initial design, which scales the write by one minus the anomaly probability, does not improve on the ungated memory. The visual-only control (event decoder and visual residual alone) ranks lower and flags more recovery frames on both sets (.150 and .348 on Reassembly).

\section{Conclusion}
\label{sec:conclusion}
HACT models bimanual procedural execution as a causal stream of structured hand transitions and scores each transition by its surprisal under the joint history of both hands.
On two power-tool procedures it has the highest AUPRC and F1 among the compared methods and the fewest recovery alarms at a validation-selected operating point, its fold models keep the highest AUPRC and F1 on another assembly order of the same product, and the ablations attribute the ranking mainly to the procedural history and the recovery behavior mainly to the interaction of the two hands.
The decoder is offline, the model needs event-level supervision and a procedure context, and transfer to another product is untested.

\vskip-0.5ex\noindent\textbf{Compliance with Ethical Standards.}
Private recordings were collected with written informed consent under institutional ethics approval; IMPACT-ego is used under its license.

\noindent\textbf{Acknowledgments.}
The project is funded by the Deutsche Forschungsgemeinschaft (DFG, German Research Foundation) -- SFB-1574 -- 471687386. This work was supported in part by the SmartAge project sponsored by the Carl Zeiss Stiftung (P2019-01-003; 2021-2026). The authors gratefully acknowledge the computing time provided on the high-performance computer HoreKa by the National High-Performance Computing Center at KIT (NHR@KIT). This center is jointly supported by the Federal Ministry of Education and Research and the Ministry of Science, Research and the Arts of Baden-W\"urttemberg, as part of the National High-Performance Computing (NHR) joint funding program (\url{https://www.nhr-verein.de/en/our-partners}). HoreKa is partly funded by the German Research Foundation (DFG).

\bibliographystyle{IEEEbib}
\bibliography{reference}

\end{document}